\documentclass[journal,twoside,web]{mainclass}
\usepackage{geometry}
\usepackage{setspace,lineno}

\usepackage{amsmath,amssymb,amsfonts}
\usepackage{amsthm}
\usepackage{mathrsfs}

\usepackage{graphicx}
\usepackage{float}
\usepackage{multirow}
\usepackage{booktabs}
\usepackage{multicol}
\usepackage{tabularx}
\usepackage{tablefootnote}
\usepackage[table,xcdraw]{xcolor}

\usepackage{algorithm}
\usepackage{algpseudocode}

\usepackage{listings}

\usepackage{cite}
\usepackage{manyfoot}

\usepackage{xcolor}
\usepackage{textcomp}
\usepackage{pifont}
\usepackage{soul}
\usepackage{tikz}

\definecolor{customgreen}{HTML}{00B050}
\definecolor{orcidgreen}{HTML}{A6CE39}
\newcommand{\cmark}{\textcolor{customgreen}{\ding{51}}} 
\newcommand{\xmark}{\textcolor{red}{\ding{55}}}

\newcommand{\orcidicon}[1]{%
  \textsuperscript{%
    \href{https://orcid.org/#1}{%
      \begin{tikzpicture}[baseline=-0.1em]
        \fill[orcidgreen] (0,0) circle (1.5ex);
        \node[white, scale=0.8, font=\bfseries\sffamily] at (0,0) {iD};
      \end{tikzpicture}%
    }%
  }%
}

\makeatletter
\let\NAT@parse\undefined
\makeatother

\usepackage[colorlinks=true, linkcolor=blue, citecolor=blue, urlcolor=blue]{hyperref}

\begin{document}

\author{
    Md. Abdur Rahman\orcidicon{0009-0004-3097-8576}\textsuperscript{1}, 
    Mohaimenul Azam Khan Raiaan\orcidicon{0009-0006-4793-5382}\textsuperscript{1}, 
    Sami Azam\orcidicon{0000-0001-7572-9750}\textsuperscript{2,*}, \\
    Asif Karim\orcidicon{0000-0001-8532-6816}\textsuperscript{2}, 
    Jemima Beissbarth\orcidicon{0000-0003-1154-230X}\textsuperscript{3}, 
    Amanda Leach\orcidicon{0000-0002-4638-8392}\textsuperscript{3}
    
    \small
	\textsuperscript{1}Department of Computer Science and Engineering, United International University, Dhaka, 1212, Bangladesh \\
    \small
	\textsuperscript{2}Faculty of Science and Technology, Charles Darwin University, Casuarina, 0909, NT, Australia \\
    \small
	\textsuperscript{3}Child Health Division, Menzies School of Health Research, Casuarina, 0909, NT, Australia \\
   	\small 
	\textsuperscript{*}Corresponding Author: sami.azam@cdu.edu.au
}







\title{WeCKD: Weakly-supervised Chained Distillation Network for Efficient Multimodal Medical Imaging}

\maketitle

\begin{abstract}
Knowledge distillation (KD) has traditionally relied on a static teacher-student framework, where a large, well-trained teacher transfers knowledge to a single student model. However, these approaches often suffer from knowledge degradation, inefficient supervision, and reliance on either a very strong teacher model or large labeled datasets. To address these, we present the first-ever \underline{We}akly-supervised \underline{C}hain-based \underline{KD} network (WeCKD) that redefines knowledge transfer through a structured sequence of interconnected models. Unlike conventional KD, it forms a progressive distillation chain, where each model not only learns from its predecessor but also refines the knowledge before passing it forward. This structured knowledge transfer further enhances feature learning and addresses the limitations of one-step KD. Each model in the chain is trained on only a fraction of the dataset and shows that effective learning can be achieved with minimal supervision. Extensive evaluation on six imaging datasets across otoscopic, microscopic, and magnetic resonance imaging modalities shows that it generalizes and outperforms existing methods. Furthermore, the proposed distillation chain resulted in cumulative accuracy gains of up to +23\% over a single backbone trained on the same limited data, which highlights its potential for real-world adoption.
\end{abstract}


\section{Introduction}
\IEEEPARstart{D}EEP learning (DL) has revolutionized the field of image classification, leading to significant advances in both natural and medical imaging. The introduction of convolutional neural networks (CNNs) has enabled models to automatically extract hierarchical features. Architectures such as ResNet, DenseNet, and EfficientNet have achieved state-of-the-art (SOTA) performance by optimizing depth, connectivity, and parameter efficiency. More recently, vision transformers (ViTs) \cite{dosovitskiy2020image} have shown competitive results by utilizing attention mechanisms \cite{yang2024dual, song2024medical, Zhou2023DistillingKF} for long-range dependencies. However, these advances come at the cost of increased computational complexity and data dependency, which makes them impractical in scenarios with limited labeled data. In medical imaging, where high-quality labeled datasets are scarce and expensive to obtain, traditional DL methods struggle to generalize effectively. This limitation has driven interest in transfer learning, semi-supervised learning, and KD as potential solutions to improve performance with constrained data \cite{han2024deep}.  

The application of DL in medical image classification has seen remarkable progress, particularly in disease diagnosis using otoscopic, microscopic, dermoscopic, MRI, and radiological imaging \cite{camalan2020otomatch, rao2024otonet, ting2023detection, kim2023toward}. For instance, several CNN-based models have been developed to classify diseases, with studies such as \cite{viscaino2020computer, afify2024insight} employing hybrid architectures that combine CNNs and recurrent networks to extract temporal characteristics. Ensemble models have also been investigated to enhance robustness \cite{zeng2021efficient, matta2024systematic}. Despite these advancements, most approaches remain heavily reliant on supervised learning paradigms that require extensive labeled datasets for optimal performance. Semi-supervised methods \cite{wang2024dual, weng2024semi} have attempted to address these issues using unlabeled data, but they often struggle with representation learning in complex medical datasets. Furthermore, transformer-based models \cite{wang2023transformer, song2024medical, Zhou2023DistillingKF} have been introduced to enhance feature extraction; however, their data-hungry nature and computational requirements limit their applicability in resource-constrained medical settings.  

KD improves model efficiency and transferability by training a smaller student model to mimic a larger teacher network \cite{hinton2015distilling}. Conventional KD methods rely on soft probability distributions to transfer knowledge, with variants such as self-distillation \cite{10233880, zhang2021self, 10605840} and multi-teacher distillation \cite{10368037, 9491090, dominguez2024melanoma}. In medical imaging, KD has been used for various applications, including cortical cataract, lung infection, and breast tumor classification using transformer-based distillation \cite{wang2023transformer, song2024medical} and whole slide image classification via graph-based KD \cite{bontempo2023graph}. Recent research has also explored hybrid KD strategies, such as CNN-ViT distillation to enhance medical image classification \cite{el2024hdkd, Zhou2023DistillingKF}, anomaly detection using student-teacher architectures \cite{liu2024anomaly, Zhou2023PullP}, and zero-shot visual sentiment prediction models \cite{moroto2023zero}. However, traditional KD approaches typically involve a single-stage knowledge transfer, which limits the extent to which knowledge can be progressively refined in multimodal domains.

Although KD has demonstrated effectiveness in model compression and transfer learning, several challenges remain unresolved. Existing single-stage KD methods often suffer from knowledge degradation, where the student model fails to retain the fine-grained representations learned by the teacher. Furthermore, static teacher-student frameworks struggle to generalize across diverse datasets, particularly in medical imaging applications with high intraclass variability \cite{ibrahim2021knowledge}. Multistage KD has been proposed to address these issues, with approaches such as unpaired multimodal KD \cite{loussaief2025adaptive, ahmad2024multi} and structural alignment-based KD \cite{10198386, sharma2024stalk} showing promising results. However, these methods are often sensitive to hyperparameter selection and require extensive tuning to achieve stable convergence. Furthermore, most existing KD methods assume that the teacher model provides optimal supervision, disregarding that certain features may be overfitted to the source domain, leading to sub-optimal student learning \cite{yin2024two}. 

To address these issues, we propose a novel weakly-supervised chain-based knowledge distillation (WeCKD) framework, particularly designed for medical image classification. Unlike conventional KD approaches, it employs a structured multistage distillation process, wherein knowledge is transferred through a sequence of models and refines the representations at each stage. This progressive learning strategy addresses knowledge degradation while enhancing generalization in low-data settings. The key contributions of this work are summarized as follows:  
\begin{itemize}
    \item A novel chain-based sequential knowledge distillation framework is introduced in which knowledge is progressively transferred through multiple intermediate models rather than a single-stage teacher-student paradigm. This structured approach reduces knowledge degradation, stabilizes learning, and allows student models to retain richer hierarchical representations.
    \item An attention-based feature refinement module is incorporated within the distillation process to ensure that the models in the distillation chain focus on the most discriminative regions learned by their predecessors.
    \item To improve model convergence and stability across datasets without the need for manual intervention, we utilize Bayesian optimization with Optuna to automatically tune key distillation hyperparameters.
    \item Comprehensive evaluations have been conducted on six real-world medical imaging datasets across three different modalities, including otoscopic, microscopic, and MRI, with limited labeled data to show the effectiveness of the proposed weakly-supervised method compared to existing supervised approaches. 
\end{itemize}

The remainder of the paper is organized as follows. Section \ref{lit_rev} reviews the related literature. Section \ref{methodology} details the proposed framework, including changes to the model architecture and the chain-based KD process. Section \ref{exp_results} presents the experimental results, including training analysis, ablation studies of model configurations, and comparisons with existing SOTA methods. Section \ref{discussion} provides an in-depth discussion of the findings and outlines directions for future research, followed by the conclusion in Section \ref{conclusion}, which summarizes the main contributions.
\vspace{-5pt}

\section{Related Works} \label{lit_rev}

\subsection{Traditional Machine Learning and Deep Learning-Based Approaches}
Early approaches to medical image analysis relied primarily on hand-made feature extraction combined with machine learning classifiers. For example, in a study, Viscaino et al. \cite{viscaino2020computer} used feature extraction techniques such as color coherence vectors, discrete cosine transforms, and filter banks to characterize abnormalities in the tympanic membrane. Their SVM classifier model achieved an average classification accuracy of 93.90\%. Similarly, Ting et al. \cite{ting2023detection} explored an alternative modality by analyzing magnitude spectrograms from microphone recordings in the ear using machine learning to detect otitis media with effusion. Their approach achieved an overall accuracy of 80.60\%, although the performance was somewhat constrained by variability in acoustic signal quality. Although these methods provided competitive classification accuracy, their reliance on manually engineered features often limited their adaptability to varying datasets and imaging conditions. 

To overcome these constraints and enable automatic feature extraction and end-to-end learning, DL has been widely adopted. For example, Mao et al. \cite{mao2022efficient} implemented deep convolutional networks, including ResNet, SENet, and EfficientNet, to classify otoscopic images and achieve better performance compared to traditional approaches that obtain an accuracy of 92.42\% with a strong precision-recall balance. Cha et al. \cite{cha2019automated} further improved classification performance by employing a CNN classifier set, where multiple models were independently trained and aggregated for final predictions, resulting in an accuracy of 93.67\%. Meanwhile, Camalan et al. \cite{camalan2020otomatch} proposed a content-based image retrieval system that converted a CNN into an image similarity lookup model, which allows retrieval of past cases instead of direct classification. However, their method achieved a lower accuracy of 82.60\%, indicating that retrieval-based methods may struggle with fine-grained diagnostic differentiation. 

Beyond purely image-based learning, alternative modalities and hybrid architectures have been explored. For instance, Sundgaard et al. \cite{sundgaard2022deep} proposed a CNN using wideband tympanometry data for otitis media detection, achieving an accuracy of 92.6\%, though it failed to distinguish between effusion and acute subtypes. Meanwhile, Viscaino et al. \cite{viscaino2021computer} introduced a hybrid CNN-LSTM framework for video-based otoscopy analysis, where deep feature extraction was coupled with temporal modeling to capture sequential variations in ear examinations. With the temporal feature learning approach, their system achieved 98.15\% accuracy in the otoscopic video analysis domain. Similarly, Sundgaard et al. \cite{sundgaard2021deep} employed deep metric learning, comparing contrastive loss, triplet loss, and multiclass N-pair loss functions to improve feature space separation to classify tympanic membrane abnormalities, achieving an accuracy of 86.00\% in otoscopic imaging.

Optimization strategies such as transfer learning and hyperparameter tuning have been explored to further refine model performance. In a study, Zeng et al. \cite{zeng2021efficient} used pretrained feature extractors through transfer learning by adapting DenseNet-BC169 and DenseNet-BC1615 to improve classification accuracy with limited training data, reaching up to 95.59\% accuracy in otoscopic imaging. Afify et al. \cite{afify2024insight}, on the other hand, employed Bayesian hyperparameter optimization to systematically search for optimal CNN configurations, which helped achieve a classification accuracy of 98.10\%. Moreover, Kim et al. \cite{kim2023toward} introduced a multimodal fusion strategy that combined CNNs with multilayer perceptrons (MLPs) using cross-attention mechanisms that allow the integration of tympanic membrane images with pure tone audiometry data for a more comprehensive diagnostic framework. Their model achieved 92.90\% accuracy and recall in otoscopic imaging. 
\begin{figure*}[ht!]
    \centering
    \includegraphics[scale = 0.08]{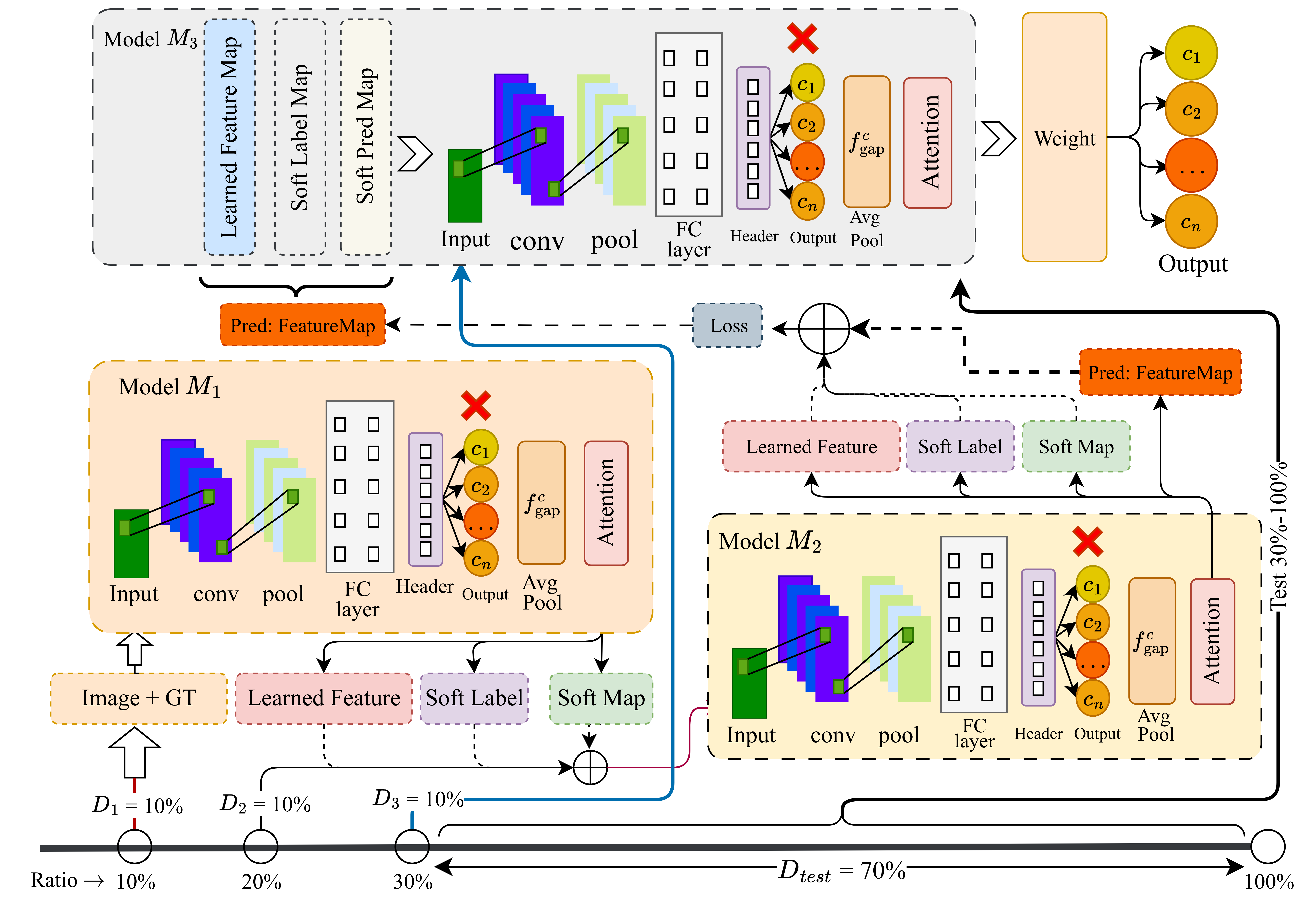}
    \caption{Illustration of the proposed weakly-supervised knowledge distillation chain. The chain forms a weakly-supervised learning process by distilling knowledge progressively across three interconnected models (\( M_1 \to M_2 \to M_3 \)), where each model learns from only 10\% of the labeled data and refines the soft predictions received from its predecessor.}
    \label{fig:WeCKD}
\end{figure*}

\vspace{-10pt}
\subsection{Knowledge Distillation-Based Approaches}
While DL has significantly improved medical image classification, model complexity often results in high computational costs and memory demands. To address such issues, KD has been widely adopted as a model compression technique, transferring knowledge from a large teacher model to a more compact student model. For example, Zhang et al. \cite{zhang2021self} introduced self-distillation, where a single network distilled its deeper representations into shallower layers, which allows intermediate classifiers to improve generalization without requiring an external teacher. Similarly, Bontempo et al. \cite{bontempo2023graph} extended KD to a graph-based multi-instance learning framework, where spatial correlations between image patches were captured using graph neural networks, and self-distillation was employed to transfer knowledge across multiple resolution scales. 

More recently, transformer-based KD techniques have been proposed to enhance feature representation. In a study, Wang et al. \cite{wang2023transformer} introduced TKD-Net, a Transformer-based KD model for cortical cataract classification, where a multimodal attention mechanism preserved clinically relevant feature relationships during knowledge transfer. However, their model achieved a relatively lower accuracy of 82.10\% in ophthalmic imaging, suggesting limitations in handling complex opacity variations. Likewise, El et al. \cite{el2024hdkd} designed a hybrid CNN-ViT KD paradigm, where a convolutional teacher model transferred knowledge to a Vision Transformer student. To improve knowledge alignment, they introduced a Mobile Channel-Spatial Attention module to address the common issue of structural mismatch between CNNs and transformers. Their approach resulted in an accuracy of 99.85\% in MRI data but showed lower performance (92.15\% accuracy; precision and recall ranging from 75.30\% to 79.80\%) in dermoscopic images. 

Alternative strategies, such as progressive distillation and anomaly detection, have been explored in a way that is beyond the standard teacher-student KD. For example, Sharma et al. \cite{sharma2024stalk} proposed StAlK, a self-KD framework where feature alignment was enforced between a mean teacher and a student model to address inter-class similarity issues in medical image classification. Their approach resulted in 93.14\% accuracy in epithelial cell classification. Meanwhile, Liu et al. \cite{liu2024anomaly} introduced a skip-connected teacher-student architecture (Skip-TS) for anomaly detection, where a randomly initialized student decoder was trained using a direct reverse KD approach to reconstruct shallow representations from a pretrained teacher encoder. Multi-teacher and zero-shot KD approaches have also been investigated in some studies \cite{loussaief2025adaptive, ahmad2024multi, weng2024zero, moroto2023zero}. Dominguez et al. \cite{dominguez2024melanoma} proposed a multi-teacher ensemble KD framework for the classification of melanoma, where the knowledge of multiple teacher models trained on distinct datasets was fused into a single student network. Similarly, Moroto et al. \cite{moroto2023zero} explored zero-shot KD for visual sentiment prediction, where implicit relationships between unseen sentiment categories were learned through a sentiment loss function.

In addition to classification tasks, KD has been used in cross-modal adaptation and segmentation. Dou et al. \cite{dou2020unpaired} applied KD in unpaired cross-modality image segmentation, where a shared CNN encoder with modality-specific normalization layers was trained using a KL-divergence-based knowledge transfer approach to bridge the domain gap between CT and MRI images. Similarly, Ibrahim et al. \cite{ibrahim2021knowledge} integrated KD into a time series forecasting model for the clinical prognosis, where an unsupervised LSTM autoencoder first extracted latent feature representations, which were then refined by a gradient-boosting student model. Song et al. \cite{song2024medical} extended KD to medical image segmentation by introducing a Res-Transformer teacher model based on U-Net and transferring hierarchical feature representations to a lightweight ResU-Net student model to maintain performance while reducing computational complexity. 

Despite advances, several limitations persist in existing frameworks, particularly with knowledge retention, architectural flexibility, and progressive learning. Traditional KD methods often rely on a single-step teacher-student transfer and result in knowledge degradation as the student struggles to retain complex hierarchical representations. Multi-teacher approaches partially address this issue, but introduce inconsistencies due to differing teacher architectures and domain shifts. Furthermore, most KD methods assume structural similarity between teacher and student networks, which restricts the adaptability across diverse model designs. Additionally, conventional KD techniques operate statically, making them lack iterative refinement strategies that allow for gradual knowledge assimilation. To address these challenges, we present the first-ever sequential distillation mechanism where knowledge is progressively transferred across a structured chain of models, even with limited training data.
\vspace{-4pt}

\section{Methodology} \label{methodology}
The objective of this study is to introduce a weakly-supervised learning framework that progressively transfers knowledge through a chain of distillation models. The proposed framework consists of three interconnected models, where each successive model refines and builds upon the knowledge acquired by its predecessor. Figure \ref{fig:WeCKD} illustrates the overall architecture of this study.

\subsection{Problem Formulation}
Let $(x_i, y_i)_{i=1}^{N}$ be the full dataset ${D}_{\text{full}}$, where \(x_i \in \mathbb{R}^{H \times W \times C}\) represents an input image, and \(y_i \in \{1, 2, ..., K\}\) is the corresponding class label, with \(K\) being the number of categories. The dataset is partitioned into three sequential training subsets, \(\mathcal{D}_1, \mathcal{D}_2, \mathcal{D}_3\), such that \(\mathcal{D}_1 \cup \mathcal{D}_2 \cup \mathcal{D}_3 \subset \mathcal{D}_{\text{full}}\) and \(|\mathcal{D}_1| = |\mathcal{D}_2| = |\mathcal{D}_3| = 0.1N\). Each of the models in the chain is trained on 10\% of the full dataset, \(\mathcal{D}_{\text{full}}\), and the remaining 70\% of the data forms the test set, denoted as \(\mathcal{D}_\text{test}\) ($0.7N$). 

The goal is to train a sequence of models, denoted as \(M_1, M_2, M_3\), in a KD chain. The first model, \(M_1\), is trained on \(\mathcal{D}_1\) and learns a function \(f_1: x \rightarrow P(y|x; \theta_1)\), where \(\theta_1\) represents the parameters of \(M_1\). The second model, \(M_2\), is trained on \(\mathcal{D}_2\) and utilizes both the ground-truth labels and the soft predictions from \(M_1\), thereby minimizing a hybrid loss function \(\mathcal{L}_{\text{KD}}^1\). The third model, \(M_3\), is trained in a similar manner, learning from \(\mathcal{D}_3\) while distilling knowledge from \(M_2\). The optimization objective is to minimize the classification loss function \(\mathcal{L}(M_3(x; \theta_3), y)\), where \(x\) represents the input image and \(y\) the ground-truth label. 

This sequential knowledge transfer framework ensures that the models use intermediate knowledge from the previous models, with only a fraction of labeled data used at each training stage. Algorithm \ref{alg:dchain} outlines the overall proposed methodology.
\begin{algorithm}[!ht]
\caption{Proposed training algorithm for the distillation chain.}
\label{alg:dchain}
\textbf{Input:} pretrained feature extractor $f_{\text{base}}$,  
training subsets $\mathcal{D}_1, \mathcal{D}_2, \mathcal{D}_3$,  
test set $\mathcal{D}_{\text{test}}$,  
learning rate $\eta$, distillation weight $\alpha$, temperature $T$, total epochs $E$.  
\begin{algorithmic}[1]
\State \textbf{init} model seq: $M_1 \gets$ $f_{\text{base}}$
\State train $M_1$ on $\mathcal{D}_1$ using supervised loss $\mathcal{L}_{\text{CE}}$
\State freeze $M_1$ and set as teacher for $M_2$
\State \textbf{init} $M_2$ from $f_{\text{base}}$, $f_{\text{gap}}^{\text{att},c} (M_1)$
\State {\color{gray}// init $M_2$ w/ with base weights + attention features of $M_1$}
\For{$e \gets 1$ to $E$}
    \For{each mini-batch $(x, y) \sim \mathcal{D}_2$}
        \State calc\_soft\_teacher\_pred: $p_T \gets M_1(x) / T$
        \State calc\_student\_pred: $p_S \gets M_2(x)$
        \State calc\_total\_loss: $\mathcal{L} \gets \alpha \mathcal{L}_{\text{CE}} + (1 - \alpha) \mathcal{L}_{\text{KD}}$
        \State {\color{gray}// supervised and distill loss for balanced optimization}
        \State update $M_2$ using optimizer with $\eta$
    \EndFor
    \State {\color{gray}// anneal temp to sharpen probability distribution}
    \State adjust\_temp: $T \gets$ schedule($T_{\max}, T_{\min}, e, E$)
\EndFor

\State freeze $M_2$ and set as teacher for $M_3$
\State \textbf{init} $M_3$ from $f_{\text{base}}$, $f_{\text{gap}}^{\text{att},c} (M_2)$
\State {\color{gray}// distill knowledge from $M_2$ to $M_3$ for the next stage}
\For{$e \gets 1$ to $E$}
    \For{each mini-batch $(x, y) \sim \mathcal{D}_3$}
        \State calc\_soft\_teacher\_pred: $p_T \gets M_2(x) / T$
        \State calc\_student\_pred: $p_S \gets M_3(x)$
        \State calc\_total\_loss: $\mathcal{L} \gets \alpha \mathcal{L}_{\text{CE}} + (1 - \alpha) \mathcal{L}_{\text{KD}}$
        \State {\color{gray}// refine previous features + minimize divergence}
        \State update $M_3$ using optimizer with $\eta$
    \EndFor
\EndFor
\end{algorithmic}
\textbf{Output:} final model $M_3$
\end{algorithm}

\subsection{Dataset} \label{dataset}
To evaluate the effectiveness of the proposed framework under limited training conditions, we employ four publicly available otoscopic imaging datasets: (1) Datos \cite{Viscaino2020}, (2) Eardrum \cite{POLAT2021}, (3) OtoMatch \cite{camalan_2019_3595567}, and (4) EardrumDS \cite{basaran2022eardrum}. These otoscopic datasets comprise images labeled for common middle and external ear conditions. To ensure generalizability across diverse classification tasks and imaging modalities, we additionally utilize two other publicly available medical imaging datasets: (1) the Brain Tumor dataset \cite{nickparvar2021brain}, which consists of MRI scans labeled across four brain tumor categories; and (2) the Lung and Colon Cancer dataset \cite{dey2022lung}, which contains microscopic histopathology images representing various cancer subtypes. Table \ref{tab:dataset_summary} summarizes the datasets (see subclass names in Table \ref{tab:class_performance}).
\begin{table}[!ht]
\small
\centering
\caption{Overview of datasets used to validate the proposed framework, including sample sizes, imaging modalities, and subclass distributions.}
\label{tab:dataset_summary}
{
\begin{tabular}{lccc}
\toprule
\textbf{Dataset} & \textbf{Size} & \textbf{Modality} & \textbf{Subclass} \\
\midrule
Datos                 & 4,400         & Otoscopic   & 4 \\
Eardrum               & 6,696         & Otoscopic   & 7 \\
OtoMatch              & 3,203         & Otoscopic   & 3 \\
EardrumDS             & 8,655         & Otoscopic   & 9 \\
Lung \& Colon Cancer  & 25,000        & Microscopic & 5 \\
Brain Tumor           & 7,023         & MRI         & 4 \\
\bottomrule
\end{tabular}
}
\end{table}

\begin{figure*}[!ht]
    \centering
    \includegraphics[scale = 0.1]{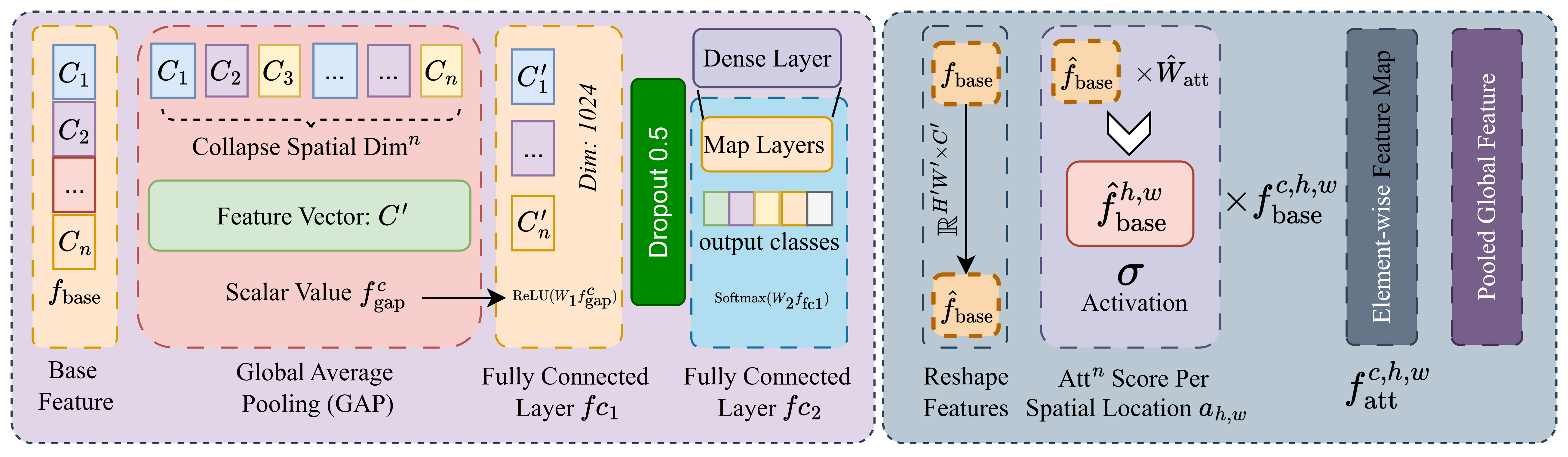}
    \caption{Proposed extension to the model: the structural inclusion and exclusion of spatial-attention and feature-refinement modules within the backbone. The extension focuses on how these components enhance the model’s focus on discriminative regions while maintaining efficiency.}
    \label{fig:base_extd}
\end{figure*}

It is worth noting that some of the datasets initially contained a limited number of images. To enhance the framework’s generalization and prevent overfitting, data augmentation techniques were applied to get the mentioned number of samples; however, no preprocessing for data quality enhancement was employed to assess the model more accurately on raw data. Additionally, the datasets are referred to by their original names as collected in their respective source directories.

\subsection{Model Architecture}
In this study, we have used a deep CNN architecture as the backbone for the feature extraction. The following sections elaborate on the base feature extraction procedure and attention mechanism used for the distillation. Figure \ref{fig:base_extd} illustrates the backbone extension. 

\subsubsection{Base Model}
The feature extraction process can be represented as a function \( f_{\text{base}}: \mathbb{R}^{H \times W \times 3} \rightarrow \mathbb{R}^{C' \times H' \times W'} \), where \( H \) and \( W \) are the input image dimensions, and \( C' \), \( H' \), and \( W' \) are the number of output channels, height, and width after applying the convolutional layers of the model. This feature map \( f_{\text{base}} \in \mathbb{R}^{C' \times H' \times W'} \) is produced by the convolutional layers, where each spatial location in the output corresponds to a learned feature representation. To aggregate the spatial information, we apply Global Average Pooling (GAP), which performs an average operation across all spatial locations \( H' \times W' \) for each of the \( C' \) channels. The GAP operation can be represented as shown in Equation \eqref{eq:GAP}:
\begin{align}
    f_{\text{gap}}^c &= \frac{1}{H' \times W'} \sum_{h=1}^{H'} \sum_{w=1}^{W'} f_{\text{base}}^{c, h, w}, \nonumber \\
    &\quad \forall c \in \{1, 2, \dots, C'\}
    \label{eq:GAP}
\end{align}
where \( f_{\text{gap}}^c \in \mathbb{R} \) represents the pooled feature for channel \( c \). The resulting feature vector is then passed through a fully connected layer, \(f_{\text{fc}}\), with 1024 neurons, represented as \( \text{ReLU}(W_1 \cdot f_{\text{gap}} + b_1) \), where \( W_1 \in \mathbb{R}^{C' \times 1024} \) and \( b_1 \in \mathbb{R}^{1024} \) are the learned weights and biases, respectively. Finally, the output from the fully connected layer, \( f_{\text{fc}} \), is passed through another dense layer with softmax activation, producing the final class probabilities. This layer applies a weight matrix \( W_2 \) and bias \( b_2 \) to transform \( f_{\text{fc}} \) into a probability distribution over \( N \) classes.

\subsubsection{Attention Mechanism}
\label{backbone_ext}
We introduced a spatial attention mechanism to enhance the model's ability to focus on the most informative regions of the input. Given the output \( f_{\text{base}} \in \mathbb{R}^{C' \times H' \times W'} \) from the backbone, we first apply a reshaping operation to flatten the spatial dimensions and treat each spatial position in the feature map as a separate feature. The flattened output \( \hat{f}_{\text{base}} \in \mathbb{R}^{H' W' \times C'} \) is then passed through a fully connected layer to compute the attention scores for each spatial location as defined in Equation \eqref{eq:att_score}:
\begin{equation}
    a_{h,w} = \sigma \left( W_{\text{att}} \cdot \hat{f}_{\text{base}}^{h,w} + b_{\text{att}} \right)
    \label{eq:att_score}
\end{equation}
where \( \hat{f}_{\text{base}}^{h,w} \in \mathbb{R}^{C'} \) is the flattened feature vector for the spatial location \((h, w)\), and \( W_{\text{att}} \in \mathbb{R}^{C' \times 1} \) is the learned weight matrix for the attention mechanism, with \( \sigma(\cdot) \) denoting the sigmoid activation function that outputs attention weights in the range \( [0, 1] \). The bias \( b_{\text{att}} \) is shared across all locations, and the output \( a_{h,w} \) represents the attention score for the spatial location \((h, w)\). The attention map \( A_{\text{att}} \in \mathbb{R}^{H' \times W'} \) is then reshaped to match the dimensions of the original feature map, \( f_{\text{base}} \), and create a spatially weighted attention map. The feature map is then weighted by the attention map, \( f_{\text{att}} \),  using an element-wise multiplication. The attention-modulated feature map \( f_{\text{att}} \) is then passed through the same fully connected layers as the base model. The GAP operation is applied to aggregate the spatial information as shown in Equation \eqref{eq:spatial_inf}:
\begin{equation}
    f_{\text{gap}}^{\text{att},c} = \frac{1}{H' \times W'} \sum_{h=1}^{H'} \sum_{w=1}^{W'} f_{\text{att}}^{c, h, w}
    \label{eq:spatial_inf}
\end{equation}
where \( f_{\text{gap}}^{\text{att}, c} \) represents the attention-enhanced feature vector for channel \( c \). The attention-weighted feature vector, \( f_{\text{gap}}^{\text{att}} \), is then processed through the fully connected layers, where a transformation is applied using learned weights and biases, followed by a ReLU activation to introduce non-linearity. The refined feature representation, \( f_{\text{fc}}^{\text{att}} \), is subsequently passed through a softmax layer, which converts it into a probability distribution over the target classes. 

\vspace{-10pt}

\subsection{Knowledge Distillation Chain}
\subsubsection{Knowledge Transfer}
\begin{figure}[ht!]
\centering
\includegraphics[scale = 0.095]{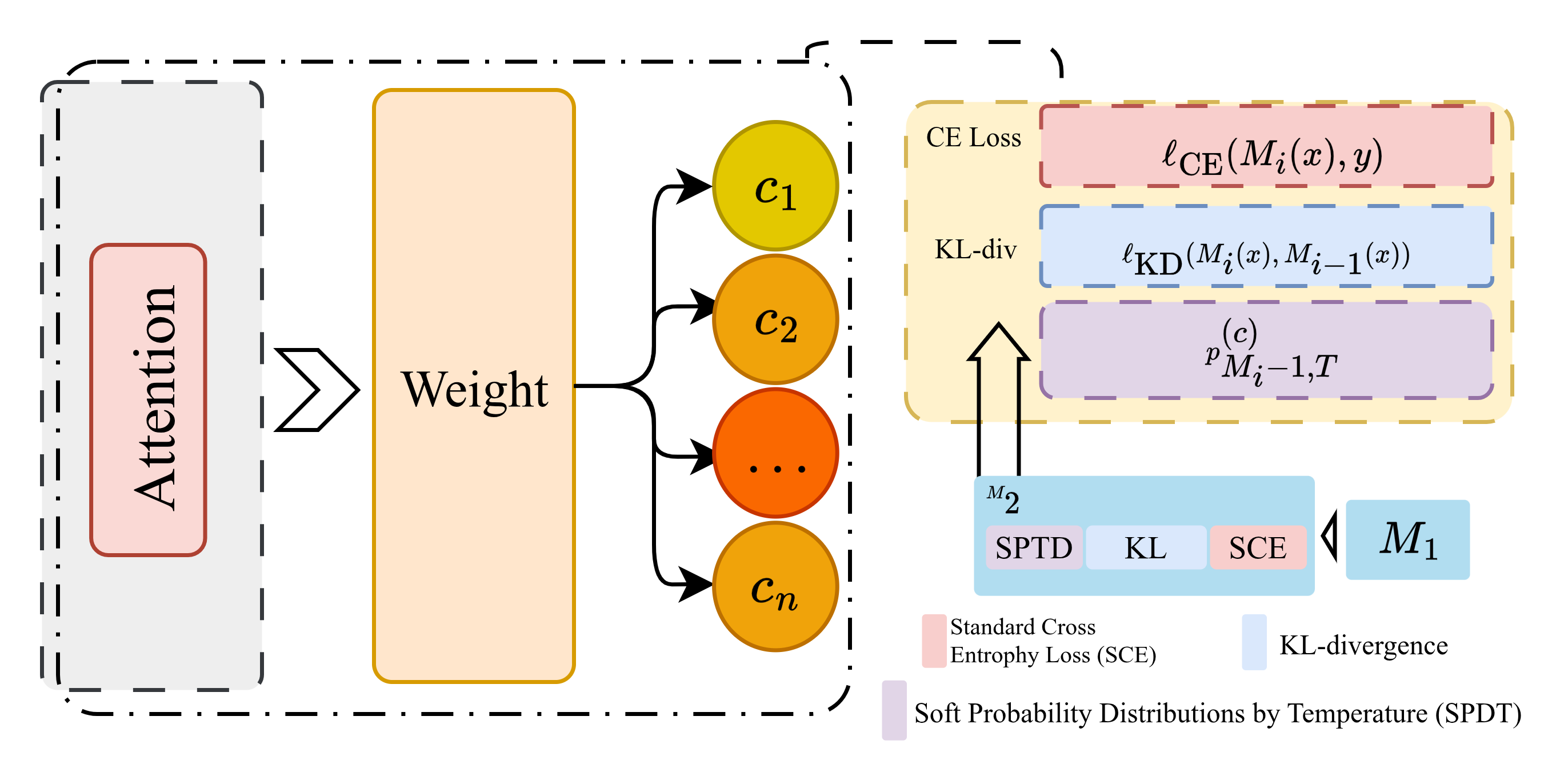}
\caption{Illustration of the multi-stage teacher–student relationships, where each subsequent model receives both hard labels and temperature-scaled soft logits from the previous stage.}
\vspace{-15pt}
\label{fig:distill}
\end{figure}

The proposed framework enables progressive knowledge transfer through a structured sequence of models, \( M_1 \), \( M_2 \), and \( M_3 \), each learning from both a newly introduced subset of the dataset and the softened predictions of its predecessor. Unlike conventional KD frameworks, which involve a single teacher-student setup, our proposed approach extends the distillation process across multiple stages and progressively refines the framework's KD and generalization ability. Figure \ref{fig:distill} briefly illustrates the distillation process. 

Formally, we express the distillation chain as a recursive mapping in which each model $(M_i)$ is obtained through the training operator \(\mathcal{F}(M_{i-1}, \mathcal{D}_i; \theta_i)\), where \(\mathcal{F}(\cdot)\) updates the model parameters based on both data-driven and teacher-guided gradients. Each model therefore learns within its own data subspace \(\mathcal{D}_i\) but inherits informative priors from \(M_{i-1}\). This recursive structure implicitly constructs a hierarchy of hypothesis spaces 
\(\mathcal{H}_1 \subseteq \mathcal{H}_2 \subseteq \mathcal{H}_3\) and enables progressive refinement under weak supervision. The weakly-supervised signifies that each stage accesses only a small labeled subset (\(|\mathcal{D}_i|/|\mathcal{D}_{\text{full}}|\!\approx\!0.1\)), yet collectively the chain learns a representation equivalent to training on the full dataset. Unlike training a single network on the aggregated data, the chain formulation promotes gradual knowledge refinement: earlier teachers capture coarse inter-class relations, while later students specialize with richer data exposure and smoother decision boundaries. To achieve effective knowledge transfer, each student model \( M_i \) minimizes a hybrid loss function that combines cross-entropy loss on hard labels from the dataset and distillation loss derived from the previous model’s soft predictions. Let \( p_{M_i} \) denote the softmax output of model \( M_i \) and \( p_{M_{i-1}} \) denote the softened probability distribution of its predecessor. The total optimization objective for each student model is shown in Equation \eqref{eq:total_loss}: 
\begin{align}
\mathcal{L}_{M_i} &= \alpha \mathbb{E}_{(x,y) \sim \mathcal{D}_i} \big[ \ell_{\text{CE}}(M_i(x), y) \big] \nonumber \\
&\quad + (1 - \alpha) \mathbb{E}_{x \sim \mathcal{D}_i} \big[ \ell_{\text{KD}}(M_i(x), M_{i-1}(x)) \big]
\label{eq:total_loss}
\end{align}
where \( \alpha \in [0,1] \) is a hyperparameter that controls the balance between direct supervision and KD. The first term in Equation \eqref{eq:total_loss} represents the standard cross-entropy loss function, defined in Equation \eqref{eq:ce_loss}: 
\begin{equation}
\ell_{\text{CE}}(M_i(x), y) = - \sum_{c=1}^{K} y_c \log p_{M_i}^{(c)}
\label{eq:ce_loss}
\end{equation}  
where \( y_c \) is the ground-truth label in a one-hot encoding form, and \( p_{M_i}^{(c)} \) is the probability predicted by model \( M_i \) for class \( c \). The second term in Equation \eqref{eq:total_loss} represents the KD loss, which minimizes the KL-divergence between the teacher model’s softened probability distribution and the student’s predicted probabilities. This is expressed as shown in Equation \eqref{eq:KL_loss}:
\begin{equation}
\ell_{\text{KD}}(M_i(x), M_{i-1}(x)) = \sum_{c=1}^{K} p_{M_{i-1}, T}^{(c)} \log \frac{p_{M_{i-1}, T}^{(c)}}{p_{M_i, T}^{(c)}}
\label{eq:KL_loss}
\end{equation}  
where \( p_{M_{i-1}, T}^{(c)} \) and \( p_{M_i, T}^{(c)} \) are the softmax probability distributions scaled by temperature \( T \), defined in Equation \eqref{eq:softmax_temp}: 
\begin{equation}
p_{M_i, T}^{(c)} = \frac{\exp(z_{M_i}^{(c)} / T)}{\sum_{j=1}^{K} \exp(z_{M_i}^{(j)} / T)}
\label{eq:softmax_temp}
\end{equation}  

The temperature \( T \) controls the smoothness of the probability distribution, where higher values encourage softer probability assignments and allow the student model to capture structured knowledge beyond hard labels. To ensure a gradual transition in the learning process, a dynamic temperature scaling strategy is applied, where the temperature, \(T(e)\), is annealed over training epochs according to Equation \eqref{eq:temp_anneal}:
\begin{equation}
T(e) = T_{\max} - (T_{\max} - T_{\min}) \times \frac{e}{E}
\label{eq:temp_anneal}
\end{equation}  
where \( e \) is the current epoch, \( E \) is the total number of epochs, and \( T_{\max} \) and \( T_{\min} \) are the initial and final temperatures. This allows the student model to initially learn from soft distributions before transitioning to sharper decision boundaries as training progresses. Additionally, the hybrid loss with temperature-scaled logits acts as a regularizer that constrains successive models within a bounded divergence region. This ensures that the learning trajectory of \(M_i\) remains close to its predecessor in function space, and effectively stabilizes convergence and limits variance propagation throughout the chain.

\subsubsection{Impact on Distillation Generalization}
The effectiveness of the distillation chain can be theoretically justified by analyzing its impact on generalization. Let \(\hat{R}(M_i)\) denote the empirical risk of model \(M_i\) on its local subset and \(R(M_i)-\hat{R}(M_i)\) represent the generalization gap $\Delta_i$. The objective of the proposed chain is to ensure that both \(R(M_i)\) and \(\Delta_i\) decrease as \(i\) increases. Each model benefits from two complementary mechanisms: (1) the progressive inclusion of data, which reduces sampling variance, and (2) the regularizing influence of distillation, which constrains consecutive models to remain close in their hypothesis space. Given a classification model \( M \) trained on dataset \( \mathcal{D} \), the expected generalization error can be given by Equation \eqref{eq:gen_error}:
\begin{equation}
R(M) = \mathbb{E}_{(x,y) \sim \mathcal{D}} [\ell(y, M(x))]
\label{eq:gen_error}
\end{equation}  
For two consecutive models \(M_{i-1}\) and \(M_i\) with predictive distributions \(p_{M_{i-1},T}\) and \(p_{M_i,T}\), standard stability and PAC–Bayes reasoning yield an approximate upper bound:
\begin{equation}
R(M_i) \le R(M_{i-1}) 
+ \mathrm{KL}\!\big(p_{M_i,T}\,\|\,p_{M_{i-1},T}\big)
+ \mathcal{O}\!\left(\tfrac{1}{\sqrt{|\mathcal{D}_i|}}\right)
\label{eq:gen_bound_x}
\end{equation}
where the first term propagates the expected performance of the teacher, the second term quantifies divergence in hypothesis space, and the final term captures finite-sample uncertainty. As the chain progresses, the data term \(|\mathcal{D}_i|\) increases while the divergence term is kept small through temperature-scaled alignment, resulting in a monotonic reduction in expected risk.

For this sequential setup, the classification risk satisfies the bound: \(R(M_3) \leq R(M_2) \leq R(M_1)\). This inequality follows naturally from the bounded–divergence assumption in Equation \eqref{eq:gen_bound_x}, which implies that each successive student achieves lower expected risk through smoother transfer and reduced empirical variance. The distillation regularizer acts as an information–theoretic smoothing term that constrains the learner to remain within a low-divergence neighborhood of its predecessor. Hence, each training stage converges toward a flatter minimum in the loss landscape, improving stability and generalization. This gradual reduction in risk arises from three key factors: (1) the progressive expansion of knowledge, where each student model has access to a larger cumulative distilled knowledge than its predecessor; (2) the regularization effect of KD, which prevents overfitting by enforcing consistency between models; and (3) the soft supervision provided by distillation, which guides the student model towards smoother and more structured decision boundaries. This tendency is reflected in the observed risk hierarchy \(R(M_3)<R(M_2)<R(M_1)\) across all tested sets

\subsubsection{Addressing Errors Accumulation}
A potential concern in multi-step KD is the accumulation of errors across models, where inaccurate predictions from earlier models may propagate through the chain. We formulate the transferred logits from model $M_{i-1}$ be expressed as the sum of $z_{M_{i-1}}$ and a perturbation term \(\delta_{i-1}\), where \(\delta_{i-1}\) denotes the perturbation induced by prediction uncertainty. If the student model \(M_i\) learns under this noisy supervision, the total propagated uncertainty after \(k\) stages can be approximated as the recursive sequence \(\delta_k \approx \beta \, \delta_{k-1}\), where \(\beta \in [0,1)\) is the attenuation factor imposed by temperature scaling and regularization. When \(\beta < 1\), the perturbation term \(\delta_k\) decays geometrically, ensuring that the cumulative error remains bounded, with the total perturbation up to step \(k\) being less than \(\delta_1\) divided by \(1 - \beta\). However, we have mitigated this issue with a controlled knowledge transfer mechanism. We defined the classification error for any model \( M_i \) as shown in Equation \eqref{eq:classification_error}:
\begin{equation}
\epsilon_i = \mathbb{E}_{(x,y) \sim \mathcal{D}_i} [1 - \mathbb{I} (\arg\max p_{M_i} = y)]
\label{eq:classification_error}
\end{equation}  
Thus, by modeling the error propagation in sequential distillation, the error rate for the final model satisfies the approximation  \(\epsilon_3 \approx \epsilon_2 + \mathcal{O}(\epsilon_2 - \epsilon_1)\); and under the bounded–divergence assumption from the previous subsection, the perturbation term \(\mathcal{O}(\epsilon_2 - \epsilon_1)\) remains within a finite tolerance governed by \(\mathrm{KL}(p_{M_2,T}\|p_{M_1,T})\). Hence, even if small biases are inherited from earlier models, the cumulative risk remains stable and non-amplifying. Since the term \( \epsilon_2 - \epsilon_1 \) is small due to the smooth knowledge transfer mechanism, the overall accumulated error remains bounded and ensures that the distilled knowledge is preserved and refined rather than degraded across the chain. The proposed dynamic temperature annealing further reduces \(\beta\) in later stages and allows \(M_3\) to emphasize reliable features and suppress early-stage misclassifications. Consequently, the chain behaves as a contractive mapping in the function space and converges to a stable hypothesis that minimizes cumulative divergence across all stages.
\begin{table*}[!ht]
\centering
\caption{Performance comparison with transfer learning backbones on 30\%. Our reported results are obtained with only the single backbone \(M_1\), without the extensions (i.e., attention mechanism and distillation chain).}
\vspace{-7pt}
\label{tab:backbone_comparison}
\fontsize{12pt}{20pt}\selectfont 
\resizebox{\textwidth}{!}{%
\begin{tabular}{lcccccccccccc}
\hline
\multirow{2}{*}{\textbf{Models}} & \multicolumn{4}{c}{\textbf{Lung and Colon Cancer (Microscopic)}} & \multicolumn{4}{c}{\textbf{OtoMatch (Otoscopic)}} & \multicolumn{4}{c}{\textbf{Brain Tumor (MRI)}} \\
\cmidrule(lr){2-5} \cmidrule(lr){6-9} \cmidrule(lr){10-13}
& Train Loss & Val Loss & Train Acc. & Val Acc. & Train Loss & Val Loss & Train Acc. & Val Acc. & Train Loss & Val Loss & Train Acc. & Val Acc. \\
\hline
ResNet50               & 0.47 & 0.55 & 72.5\% & 69.3\% & 0.45 & 0.52 & 74.2\% & 70.8\% & 0.53 & 0.60 & 65.3\% & 60.5\% \\
ResNet101              & 0.50 & 0.58 & 70.8\% & 67.5\% & 0.48 & 0.56 & 72.0\% & 69.0\% & 0.55 & 0.63 & 64.0\% & 58.7\% \\
VGG16                  & 0.52 & 0.60 & 68.9\% & 65.1\% & 0.50 & 0.57 & 70.5\% & 66.7\% & 0.58 & 0.66 & 59.4\% & 54.2\% \\
VGG19                  & 0.51 & 0.59 & 69.4\% & 66.0\% & 0.49 & 0.56 & 71.0\% & 67.8\% & 0.57 & 0.65 & 60.0\% & 55.3\% \\
DenseNet121            & 0.45 & 0.53 & 74.1\% & 70.5\% & 0.42 & 0.50 & 76.3\% & 72.9\% & 0.50 & 0.58 & 67.2\% & 62.4\% \\
DenseNet201            & 0.46 & 0.54 & 73.5\% & 70.0\% & 0.43 & 0.51 & 75.8\% & 72.4\% & 0.51 & 0.59 & 66.8\% & 61.7\% \\
Xception               & 0.44 & 0.52 & 75.0\% & 72.1\% & 0.41 & 0.49 & 77.4\% & 74.0\% & 0.49 & 0.56 & 69.0\% & 64.8\% \\
EfficientNet-B0        & 0.43 & 0.51 & 76.2\% & 73.5\% & 0.40 & 0.48 & 78.8\% & 75.2\% & 0.48 & 0.54 & 70.3\% & 66.5\% \\
\rowcolor{gray!20}Ours & 0.34 & 0.37 & 78.3\% & 76.2\% & 0.32 & 0.36 & 80.1\% & 77.5\% & 0.72 & 0.75 & 71.5\% & 68.9\% \\
\hline
\end{tabular}%
}
\end{table*}
The structured nature allows each model to benefit from incremental data exposure and progressively learned feature representations, which makes the final model \( M_3 \) significantly more resilient and generalizable than its predecessors. This bounded-error characteristic also guarantees convergence of the chain toward a fixed point \(M^*\) that satisfies \(M^*\) defined by \(\mathcal{F}(M^*, \mathcal{D})\) and ensures long-term stability even under weak supervision.

\vspace{-8pt}
\section{Experimental Results Analysis}     
\label{exp_results}
\begin{table*}[ht!]
\centering
\caption{Progressive improvement across the distillation chain over six datasets. Accuracies and corresponding error reductions are reported at each stage.}
\vspace{-7pt}
\label{tab:kd_chain_progression}
\renewcommand{\arraystretch}{1.2}
\resizebox{\textwidth}{!}{%
\begin{tabular}{lcccccc|cccccc}
\toprule
\multirow{2}{*}{\textbf{Dataset}} & 
\multicolumn{6}{c}{\textbf{Accuracy Over the Chain $\uparrow$}} & 
\multicolumn{6}{c}{\textbf{Loss Over the Chain $\downarrow$}} \\
\cmidrule(lr){2-7} \cmidrule(lr){8-13}
& $\mathbf{M_1}$ & $\mathbf{M_1 \to M_2}$ & $\mathbf{\Delta(M_1 \to M_2)}$ 
& $\mathbf{M_2 \to M_3}$ & $\mathbf{\Delta(M_2 \to M_3)}$ & $\mathbf{\Delta(M_1 \to M_3)}$ 
& $\mathbf{M_1}$ & $\mathbf{M_1 \to M_2}$ & $\mathbf{\Delta(M_1 \to M_2)}$ 
& $\mathbf{M_2 \to M_3}$ & $\mathbf{\Delta(M_2 \to M_3)}$ & $\mathbf{\Delta(M_1 \to M_3)}$ \\
\midrule
Datos                & 70.35\% & 78.72\% & +8.37\%  & 93.30\% & +14.58\% & +22.95\% &  0.4172 & 0.3522 & 0.0650 & 0.2263 & 0.1259 & 0.1909 \\
Eardrum              & 69.50\% & 82.07\% & +12.57\% & 97.95\% & +15.88\% & +28.45\% &  0.1817 & 0.1019 & 0.0798 & 0.0945 & 0.0074 & 0.0872 \\
OtoMatch             & 73.11\% & 78.41\% & +5.30\%  & 89.51\% & +11.10\% & +16.40\% &  0.3641 & 0.3128 & 0.0513 & 0.2776 & 0.0352 & 0.0865 \\
EardrumDS            & 74.75\% & 83.01\% & +8.26\%  & 96.00\% & +12.99\% & +21.25\% &  0.2890 & 0.2519 & 0.0371 & 0.1975 & 0.0544 & 0.0915 \\
Brain Tumor          & 64.14\% & 74.31\% & +10.17\% & 89.36\% & +15.05\% & +25.22\% &  0.7522 & 0.7064 & 0.0458 & 0.6111 & 0.0953 & 0.1411 \\
Lung \& Colon Cancer & 71.15\% & 82.22\% & +11.07\% & 97.40\% & +15.18\% & +26.25\% &  0.3782 & 0.3005 & 0.0777 & 0.1519 & 0.1486 & 0.2263 \\
\bottomrule
\end{tabular}
}
\vspace{-15pt}
\end{table*}
\subsection{Training Procedure}
We conducted the training using mini-batch gradient descent with a batch size of 32, and all images are resized and scaled to \( 224\times224 \) pixels for consistency. The initial learning rate is set to \( \eta = 10^{-3} \) and follows a decay schedule, reducing by a factor of 0.1 when validation loss stagnates. Each model is trained for up to 50 epochs, with early stopping applied if validation loss does not improve for 10 consecutive epochs. The hyperparameter optimization and selection process is discussed in Section \ref{ablation}. 

\vspace{-15pt}
\subsection{Backbones for Feature Extraction}
To determine the most suitable feature extraction backbone, we conducted an initial experiment with several SOTA DL architectures using standard cross-entropy loss without the distillation chain or additional modifications. In this phase (as shown in Table \ref{tab:backbone_comparison}), we selected a separate set comprising 30\% of the data and trained models including ResNet50, ResNet101, VGG16, VGG19, DenseNet121, DenseNet201, Xception, and EfficientNet-B0. Each model was independently trained on three datasets: Lung and Colon Cancer, OtoMatch, and Brain Tumor, representing the distinct imaging modalities of microscopic, otoscopic, and MRI, respectively.

The results indicate that in low-data training scenarios, the proposed model consistently outperforms other deeper architectures, such as ResNet101 and DenseNet201. This also suggests that deeper networks require significantly more training data to generalize effectively and may not be optimal when data availability is limited. While EfficientNet-B0 and Xception demonstrated competitive performance, our model provided a better trade-off between model complexity and generalization and achieved the highest accuracy with the lowest validation loss in all datasets. Following these findings, we extended the backbone's architecture with attention modules and the KD chain. 

\subsection{Distillation Chain Evaluation} \label{chain_eval}
To evaluate the effectiveness of the proposed sequential distillation strategy, we conducted a comprehensive analysis across the entire chain and examined how each successive model benefits from the knowledge transferred by its predecessor and whether the chain improves feature generalization and reduces training loss on limited, weakly-supervised data.

As seen in Table \ref{tab:kd_chain_progression}, the training accuracy exhibits a gradual increase from \( M_1 \) to \( M_3 \) across all 3 modalities, accompanied by a complementary reduction in training loss. On average, \( M_2 \) improves accuracy by approximately \( +9.29\% \) over \( M_1 \), while the final model \( M_3 \) achieves a further gain of \( +14.13\% \), leading to a total improvement of \( +23.42\% \) through the chain. This pattern validates that the intermediate student does not merely replicate the teacher’s predictions but internalizes softened representations that facilitate more stable optimization in the subsequent stage. On the other hand, the corresponding loss values also decrease (avg. 0.3970 in \( M_1 \) to 0.2598 in \( M_3 \)) in proportion to the accuracy gains, which also confirms better convergence and reduced gradient noise. 
\begin{figure*}[ht!]
\centering
\includegraphics[scale = 0.3]{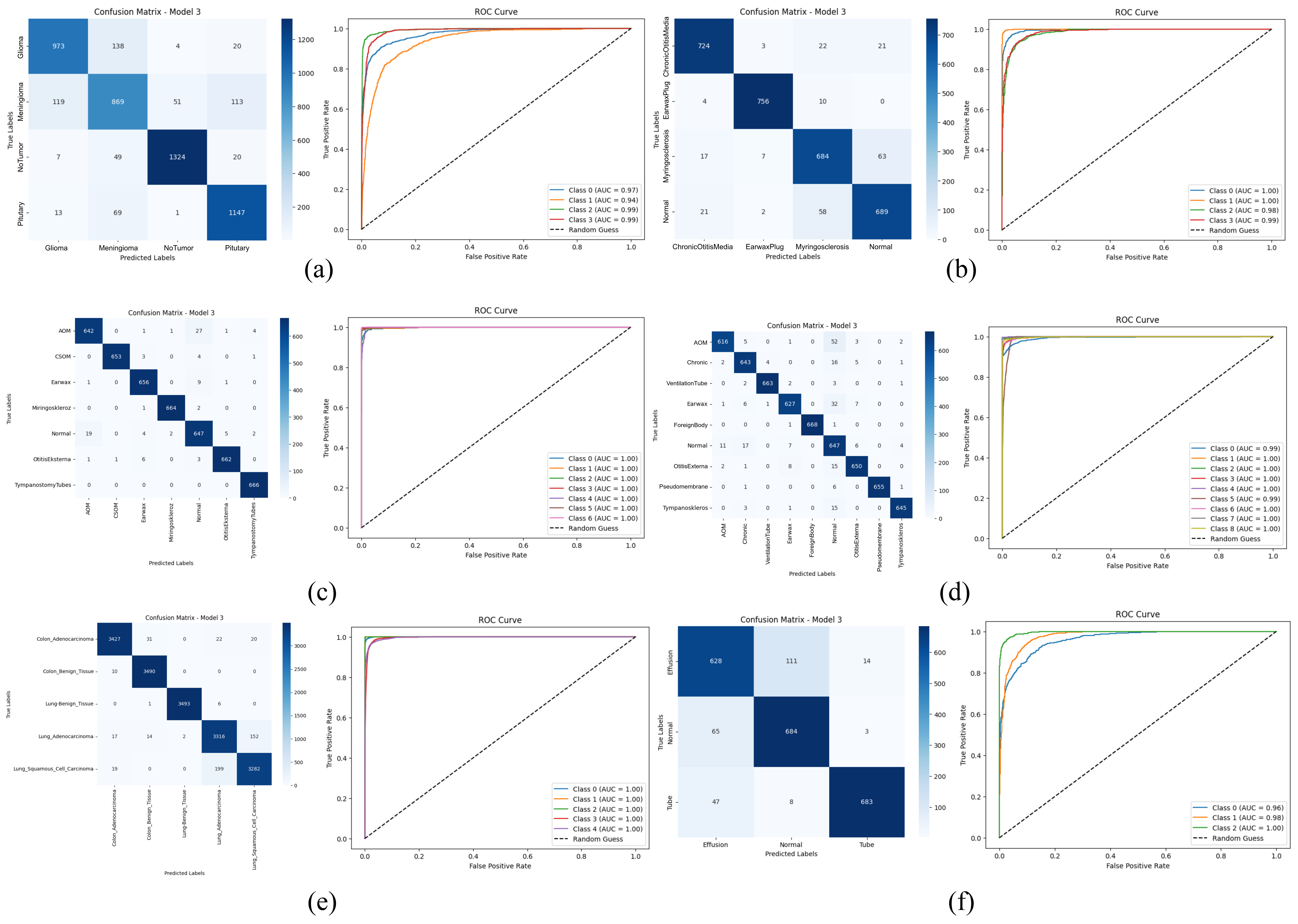}
\caption{Visualization of the confusion matrices and AUC-ROC curves of the final distilled model \(M_3\) for the six medical imaging datasets: (a) Brain Tumor; (b) Datos; (c) Eardrum; (d) EardrumDS; (e) Lung and Colon Cancer; and (f) Otomatch.}
\label{fig:conf_mat}
\vspace{-15pt}
\end{figure*}

More importantly, the results suggest that a single backbone trained on 30\% of the dataset (as seen in Table \ref{tab:backbone_comparison}) is decisively outperformed by the proposed distillation chain (i.e., \( M_1 \to M_2 \to M_3 \)) under weakly-supervised conditions of the same 30\% data. This improvement arises because the chained framework incrementally preserves and refines learned distributions and allows each subsequent model to operate on a smoother feature manifold inherited from its predecessor. As a result, instead of relearning from the allocated data (i.e., 10\% each), each model uses the softened inter-class boundaries and calibrates logits transferred from the previous stage.

Following this progression, the final model \( M_3 \) (with the backbone extension) achieves training accuracies across datasets ranging from 89.36\% to 97.95\%, with EardrumDS achieving the highest accuracy (97.95\%) and Brain Tumor the lowest (89.36\%). On the other hand, EardrumD achieved the lowest training loss of 0.0945, while Brain Tumor has the highest loss (0.6111). 

Figure \ref{fig:conf_mat} illustrates the confusion matrices and AUC-ROC curves of \( M_3 \), and Table \ref{tab:inf_time} reports the steps taken by the model per epoch with the time (in ms) for each step.
\vspace{-7pt}
\begin{table}[ht!]
\centering
\small
\caption{Training-time characteristics of the final distilled model. Step counts per epoch and corresponding processing times (in ms) are listed for each dataset.}
\vspace{-7pt}
\label{tab:inf_time}
\small
\renewcommand{\arraystretch}{1.2}
\begin{tabular}{lcc}
\toprule
\textbf{Dataset} &\textbf{Step/Epoch} & \textbf{Time/Step} \\
\midrule
Datos                & 13  & 114 ms \\
Eardrum              & 20  & 124 ms \\
OtoMatch             & 10  & 200 ms \\
EardrumDS            & 27  & 137 ms \\
Brain Tumor          & 21  & 128 ms \\
Lung \& Colon Cancer & 78  & 217 ms \\
\bottomrule
\end{tabular}
\end{table}

\vspace{-10pt}
\subsection{Individual Class Performance Analysis}
\label{subclass_eval}
Table \ref{tab:overall_performance} reports the classification performance of the proposed framework on the test set of the remaining 70\% data. The highest accuracies are observed in the Eardrum and Lung and Colon Cancer datasets, with accuracies of 97.88\% and 97.18\%, respectively. In contrast, the Brain Tumor dataset shows a lower overall accuracy of 87.70\%.

\vspace{-5pt}
\begin{table}[!ht]
\centering
\caption{Summary of the model’s classification performance on the (70\%) test datasets.}
\label{tab:overall_performance}
\small
\vspace{-7pt}
\begin{tabular}{lcccc}
\toprule
\textbf{Dataset} & \textbf{Acc} & \textbf{Prec} & \textbf{Recall} & \textbf{F1-score} \\
\midrule
Datos                  & 92.61\%  & 92.43\%  & 92.59\%  & 92.51\%  \\
Eardrum                & 97.88\%  & 97.91\%  & 97.90\%  & 97.90\%  \\
OtoMatch               & 88.92\%  & 89.15\%  & 88.94\%  & 88.99\%  \\
EardrumDS              & 95.94\%  & 96.18\%  & 95.94\%  & 96.00\%  \\
Brain Tumor            & 87.70\%  & 87.68\%  & 87.71\%  & 87.70\%  \\ 
Lung \& Colon Cancer   & 97.18\%  & 97.18\%  & 97.18\%  & 97.18\%   \\ 
\bottomrule
\end{tabular}
\end{table}

\begin{figure*}[ht!]
    \centering
    \includegraphics[scale = 0.08]{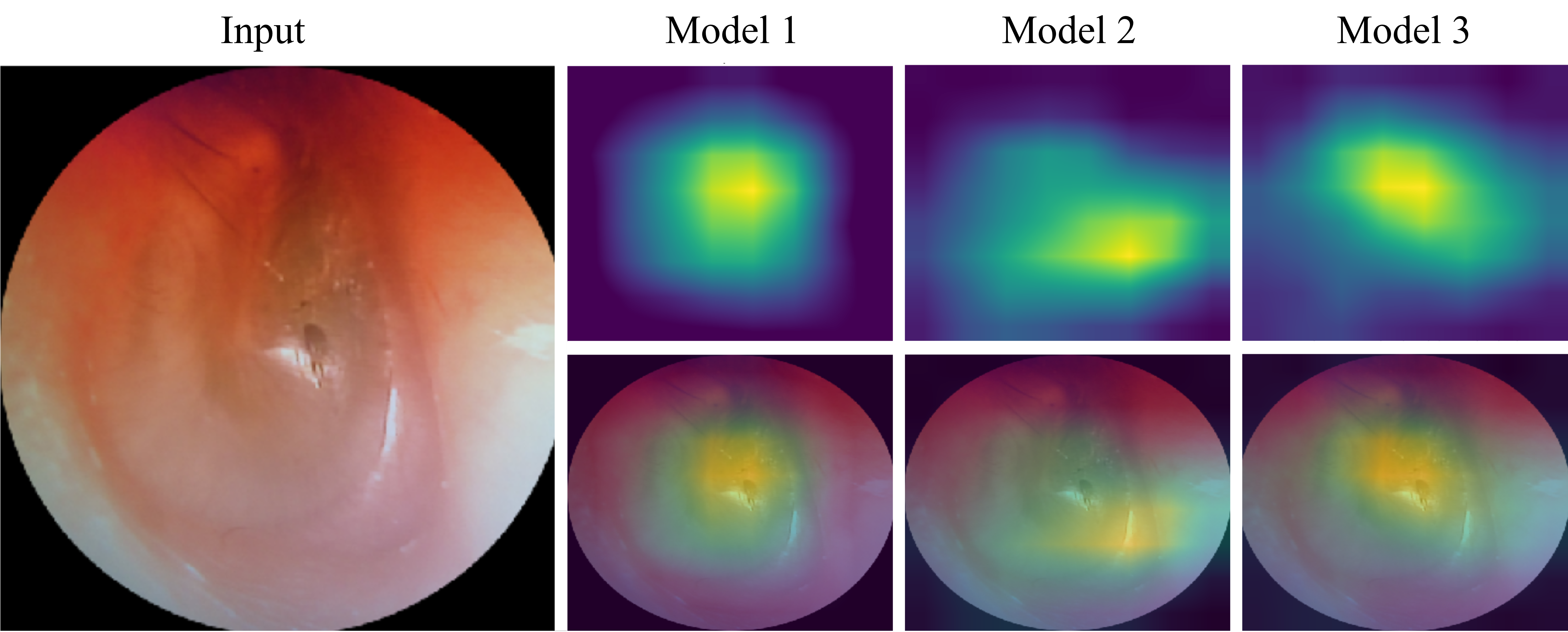}
    \caption{Gradient-weighted class activation mapping visualization of an otoscopic image. The visualization highlights the spatial regions that most influence the model’s predictions, and it shows that the attention mechanism effectively aligns these activations with clinically relevant structures in the tympanic membrane.}
    \label{fig:gradcam}
    \vspace{-10pt}
\end{figure*}

\begin{table}[!ht]
\small
\centering
\caption{Individual class performance across datasets. \textbf{(L)}, and \textbf{(C)} in the last five rows represent \textbf{L}ung, and \textbf{C}olon, respectively.}
\vspace{-7pt}
\label{tab:class_performance}
\fontsize{7.5pt}{9pt}\selectfont 
\renewcommand{\arraystretch}{1.2}{
\begin{tabular}{clccc}
\toprule
\textbf{Dataset} & \textbf{Subclass} & \textbf{Precision} & \textbf{Recall} & \textbf{F1-score} \\
\midrule
\multirow{4}{*}{\rotatebox{90}{Datos}} 
& Earwax Plug             & 98.43\%    & 98.18\% & 98.30\%   \\
& Chronic Otitis Media    & 94.51\%    & 94.02\% & 94.26\%   \\
& Myringosclerosis        & 88.37\%    & 88.71\% & 88.54\%   \\
& Normal                  & 88.44\%    & 89.48\% & 88.95\%   \\
\midrule
\multirow{7}{*}{\rotatebox{90}{Eardrum}} 
& Otitis Eksterna         & 98.95\%    & 98.36\% & 98.65\%   \\
& Miringoskleroz          & 99.55\%    & 99.55\% & 99.55\%   \\
& Tympanostomy Tubes      & 98.96\%    & 100.0\% & 99.48\%   \\
& CSOM                    & 99.84\%    & 98.79\% & 99.31\%   \\
& Earwax                  & 97.76\%    & 98.35\% & 98.05\%   \\
& Normal                  & 93.49\%    & 95.28\% & 94.38\%   \\
& AOM                     & 96.83\%    & 94.97\% & 95.89\%   \\
\midrule
\multirow{3}{*}{\rotatebox{90}{OtoMatch}} 
& Effusion                & 84.87\%    & 83.40\% & 84.13\%   \\
& Normal                  & 85.18\%    & 90.96\% & 87.97\%   \\
& Tube                    & 97.57\%    & 92.55\% & 94.99\%   \\
\midrule
\multirow{9}{*}{\rotatebox{90}{EardrumDS}}
& Otitis Externa & 96.87\% & 96.15\% & 96.50\% \\
& Foreign Body & 100.0\% & 99.70\% &  99.85\% \\
& Earwax & 96.90\% & 93.03\% & 94.92\% \\
& Pseudomembrane & 100.0\% & 98.70\% & 99.38\% \\
& Chronic & 94.84\% & 95.83\% & 95.33\% \\
& Normal & 82.81\% & 93.49\% & 87.83\% \\
& AOM & 97.47\% & 90.72\% & 93.97\% \\
& Ventilation Tube & 99.25\% & 98.80\% & 99.02\% \\
& Tympanoskleros & 98.62\% & 97.14\% & 97.87\% \\
\midrule
\multirow{4}{*}{\rotatebox{90}{Brain Tumor}} 
& Pituitary               & 88.23\%    & 93.25\% & 90.67\%   \\
& No Tumor                & 95.94\%    & 94.57\% & 95.25\%   \\
& Meningioma              & 72.24\%    & 75.43\% & 73.80\%   \\
& Glioma                  & 87.50\%    & 85.73\% & 86.60\%   \\
\midrule
\multirow{5}{*}{\rotatebox{90}{\shortstack{(L)ung \&\\ (C)olon Cancer}}}
& (L) Adenocarcinoma      & 93.59\%    & 94.71\% & 94.14\%   \\
& (C) Benign Tissue       & 98.69\%    & 99.71\% & 99.19\%   \\
& (L) Benign Tissue       & 99.94\%    & 99.80\% & 99.87\%   \\
& (C) Adenocarcinoma      & 98.67\%    & 97.31\% & 97.98\%   \\
& (L) Squamous Cell Carcinoma  & 95.02\%    & 97.71\% & 96.34\%   \\
\bottomrule
\end{tabular}
}
\vspace{-15pt}
\end{table}

To further interpret the performance at a finer granularity, we explored individual class metrics across all datasets. The complete results, as detailed in Table \ref{tab:class_performance}, confirm that WeCKD effectively captures complex feature representations and achieves high performance in multimodal classification settings while improving generalization across datasets with varying levels of class separability. The highest classification performance is observed for Miringoskleroz and Tympanostomy Tubes in the Eardrum dataset, both exceeding 99.55\% in the F1-score (see Figure \ref{fig:gradcam}). 
\vspace{-10pt}

\subsection{Ablation Experiments} \label{ablation}
Instead of manually testing different configurations, we employed Optuna \cite{akiba2019optuna} for hyperparameter optimization, which efficiently explores the search space using Tree-structured Parzen Estimators (TPE). Optuna dynamically selects promising hyperparameters based on previous evaluations, reducing the computational cost of exhaustive grid searches.

In our framework, we focus on optimizing three key hyperparameters in five trials: a) the learning rate, LR ($\eta$), which determines the step size for gradient updates; b) the distillation weight, DW ($\alpha$), which balances the trade-off between supervised loss and KD loss; and c) the temperature scaling factor, TSF ($T$), which controls the smoothness of the probability distribution of the teacher model. We defined the search space for hyperparameters as $\eta \in [10^{-5}, 10^{-2}]$, $\alpha \in [0.5, 0.9]$, and $T \in [1.0, 5.0]$. The objective function was designed to maximize the accuracy of the validation while ensuring a stable generalization. Table \ref{tab:hyperparam_tuning} presents the five detailed hyperparameter tuning trials in the datasets.
\begin{table*}[]
\centering
\caption{Optimal hyperparameter tuning results using Optuna across the datasets after five trials: Values in the best trials are highlighted. LR ($\eta$), DW ($\alpha$), and TSF ($T$) in the \textit{Aspect} column represent the learning rate, distillation weight, and temperature scaling factor, respectively.}
\vspace{-7pt}
\label{tab:hyperparam_tuning}
\scriptsize
\resizebox{\textwidth}{!}{
\begin{tabular}{cccccccc}
\toprule
\multirow{2}{*}{\textbf{Aspect}} & \multirow{2}{*}{\textbf{Trial No.}} & \multicolumn{6}{c}{\textbf{Datasets}} \\
\cmidrule(lr){3-8}
& & \textbf{Datos} & \textbf{OtoMatch} & \textbf{Eardrum} & \textbf{EardrumDS} & \textbf{Brain Tumor} & \textbf{Lung \& Colon Cancer} \\
\midrule
LR ($\eta$)
    & 0  & 0.0011  & \cellcolor{gray!20}$5.476\times10^{-5}$  & $1.62\times10^{-5}$  & 0.0002  & 0.0007  & \cellcolor{gray!20}0.0005  \\
    & 1  & 0.0059  & $1.450\times10^{-5}$  & 0.0004  & 0.0004  & 0.0001  & 0.0007  \\
    & 2  & \cellcolor{gray!20}$9.97\times10^{-5}$  & 0.0088  & 0.0016  & \cellcolor{gray!20}0.0003  & 0.0071  & 0.0009  \\
    & 3  & $7.31\times10^{-5}$  & $1.449\times10^{-5}$  & $1.05\times10^{-5}$  & 0.0002  & 0.0001  & 0.0001  \\
    & 4  & 0.0050  & $8.130\times10^{-5}$  & \cellcolor{gray!20}0.0064  & $5.176\times10^{-5}$  & \cellcolor{gray!20}0.0003  & 0.0031  \\
DW ($\alpha$)
    & 0  & 0.8209  & \cellcolor{gray!20}0.6083  & 0.8431  & 0.7830  & 0.7006  & \cellcolor{gray!20}0.7365  \\
    & 1  & 0.7588  & 0.7619  & 0.5554  & 0.7093  & 0.5826  & 0.8896  \\
    & 2  & \cellcolor{gray!20}0.7555  & 0.7573  & 0.7920  & \cellcolor{gray!20}0.6250  & 0.6317  & 0.7606  \\
    & 3  & 0.8153  & 0.6660  & 0.6750  & 0.6122  & 0.7692  & 0.7414  \\
    & 4  & 0.8601  & 0.7080  & \cellcolor{gray!20}0.8555  & 0.7750  & \cellcolor{gray!20}0.7450  & 0.6291  \\
TSF ($T$)
    & 0  & 1.553  & \cellcolor{gray!20}1.846  & 1.586  & 4.231  & 4.661  & \cellcolor{gray!20}4.358  \\
    & 1  & 3.821  & 4.547  & 3.586  & 3.209  & 3.964  & 1.421  \\
    & 2  & \cellcolor{gray!20}4.787  & 2.929  & 1.917  & \cellcolor{gray!20}4.166  & 2.808  & 1.187  \\
    & 3  & 3.060  & 4.980  & 2.464  & 1.598  & 4.104  & 1.957  \\
    & 4  & 3.908  & 4.846  & \cellcolor{gray!20}1.219  & 3.603  & \cellcolor{gray!20}2.023  & 1.853  \\
\bottomrule
\end{tabular}
}
\vspace{-2pt}
\end{table*}

\begin{table*}[!ht]
\centering
\caption{Comparison of the proposed framework with existing SOTA methods. The \textbf{Un/Weakly} column indicates whether the respective study uses an unsupervised or weakly-supervised methodology (with \xmark denoted as neither). The respective datasets in our experiments are: a) Datos; b) Eardrum; c) OtoMatch; d) EardrumDS; e) Brain Tumor; and f) Lung and Colon Cancer.}
\vspace{-5pt}
\label{tab:comparison_sota}
\small
\begin{tabular}{ccclcccc}
\toprule
\textbf{Ref} & \textbf{Knowledge Distillation} & \textbf{Un/Weakly} & \textbf{Modality} & \textbf{Accuracy} & \textbf{Precision} & \textbf{Recall} & \textbf{F1-score} \\
\midrule
\cite{sharma2024stalk} & \cmark & \xmark & Epithelial & 93.14\%  & 92.27\%  & 92.65\%  & 92.46\%  \\
\cite{el2024hdkd} & \cmark & \xmark & Dermoscopic & 92.15\%  & 75.30\%  & 79.80\%  & 77.48\%  \\
\cite{sharma2024stalk} & \cmark & \xmark & Dermoscopic & 90.46\%  & 84.87\%  & 88.19\%  & 86.50\%  \\
\cite{mao2022efficient} & \xmark & \xmark & Otoscopic & 92.42\%  & 90.98\%  & 90.31\%  & 90.64\%  \\
\cite{viscaino2020computer} & \xmark & \xmark & Otoscopic & 93.90\%  & 87.70\%  & -        & -        \\
\cite{cha2019automated} & \xmark & \xmark & Otoscopic & 93.67\%  & -        & -        & -        \\
\cite{afify2024insight} & \xmark & \xmark & Otoscopic & 98.10\%  & 98.10\%  & 98.11\%  & 98.10\%  \\
\cite{camalan2020otomatch} & \xmark & \xmark & Otoscopic & 82.60\%  & -        & -        & -        \\
\cite{viscaino2021computer} & \xmark & \xmark & Otoscopic & 98.15\%  & 91.94\%  & 91.67\%  & 91.51\%  \\
\cite{kim2023toward} & \xmark & \xmark & Otoscopic & 92.90\%  & -        & 90.90\%  & -        \\
\cite{rao2024otonet} & \xmark & \xmark  & Otoscopic & 99.30\%  & 99.50\%  & 99.30\%  & 99.40\%  \\
\cite{sundgaard2021deep} & \xmark & \xmark & Otoscopic & 86.00\%  & 85.00\%  & 79.66\%  & 82.24\%  \\
\cite{zeng2021efficient} & \xmark & \xmark & Otoscopic & 95.59\%  & -        & -        & -        \\
\cite{wang2023transformer} & \cmark & \xmark & Ophthalmic & 82.10\%  & 73.00\%  & 71.40\%  & 71.80\%  \\
\cite{el2024hdkd} & \cmark & \xmark & MRI & 99.85\%  & 99.90\%  & 99.80\%  & 99.85\%  \\
\cite{sundgaard2022deep} & \xmark & \xmark & WBT & 92.60\%  & 93.00\%  & 92.20\%  & 92.60\%  \\
\cite{ting2023detection} & \xmark & \xmark & Signals & 80.60\%  & 73.68\%  & 66.60\%  & 69.96\%  \\
\midrule
\multirow{6}{*}{\shortstack{Proposed \\ (WeCKD)}} &
\multirow{6}{*}{\cmark} &
\multirow{6}{*}{\cmark} &
a) Otoscopic     & 92.61\% & 92.43\% & 92.59\% & 92.51\% \\
& & & b) Otoscopic & 97.88\% & 97.91\% & 97.90\% & 97.90\% \\
& & & c) Otoscopic & 88.92\% & 89.15\% & 88.94\% & 88.99\% \\
& & & d) Otoscopic & 95.94\% & 96.18\% & 95.94\% & 96.00\% \\
& & & e) MRI       & 87.70\% & 87.68\% & 87.71\% & 87.67\% \\
& & & f) Microscopic & 97.18\% & 97.18\% & 97.18\% & 97.18\% \\
\bottomrule
\end{tabular}
\vspace{-15pt}
\end{table*}

Our experiments show that the optimal learning rates varied between datasets. For example, lower values, such as $\eta = 9.97 \times 10^{-5}$ for the Datos data set, stabilized training and mitigated overfitting in cases where the dataset contained overlapping class distributions. In contrast, datasets with clearer feature separability, such as Eardrum, benefited from a higher learning rate of $\eta = 0.0064$. The distillation weight $\alpha$ showed a strong correlation with the complexity of the dataset. Datasets with well-defined class boundaries, such as Eardrum ($\alpha = 0.8555$), performed optimally when the supervised loss was given a higher weight. In contrast, for datasets with greater intraclass variance, such as OtoMatch ($\alpha = 0.6083$), a lower $\alpha$ proved beneficial. The temperature scaling factor $T$ influenced the smoothness of knowledge transfer between models. Higher temperatures, such as $T = 4.787$ for the Datos dataset, allowed for better generalization by preventing overconfidence in predictions. In contrast, lower temperatures, such as $T = 1.219$ for Eardrum, proved effective in datasets with distinct class separability.
\vspace{-10pt}

\subsection{Comparison with Existing Methods}
Table \ref{tab:comparison_sota} compares WeCKD with existing SOTA DL models for medical image classification. We observe that fully supervised CNN-based approaches, such as OTONet \cite{rao2024otonet} and ensemble methods \cite{afify2024insight}, achieve greater accuracy than 98\% on otoscopic datasets, but rely on extensive labeled data. KD techniques have been explored for medical imaging \cite{wang2023transformer, el2024hdkd}, with transformer-based KD models achieving high performance but still requiring a fully trained teacher. Our model overcomes these constraints by using a structured multistage distillation chain and progressively refines knowledge using only 30\% of labeled data. Despite this constraint, it achieves 97.88\% accuracy in Eardrum and 95.94\% in EardrumDS and performs similarly to fully supervised models trained for only 50 epochs. Furthermore, the model generalizes well in MRI and microscopic datasets, achieving 87.70\% on the Brain Tumor dataset and 97.18\% on the Lung and Colon Cancer dataset. Notably, these results were obtained without additional preprocessing or enhancement of image quality, which further highlights the efficiency of our weakly-supervised distillation chain framework.

\vspace{-8pt}
\section{Discussion} \label{discussion}
Unlike conventional fully supervised learning approaches, the proposed framework progressively transfers knowledge through a structured sequence of interconnected models and allows learning from only 10\% of labeled data sequentially (for each model) in the chain while maintaining strong classification performance. The training process follows a multistage KD strategy, in which each model in the sequence refines its learned representations using both limited direct supervision and softened predictions from its predecessor. The framework ensures a stable and effective knowledge transfer mechanism while mitigating the risk of premature convergence or knowledge degradation across the chain by integrating temperature-scaled distillation and hybrid loss optimization. The structured training methodology enables it to achieve high generalization performance despite operating under weak supervision. 

A key strength of the model lies in its progressive knowledge-transfer mechanism, which allows each successive model to enhance feature representations learned by its predecessor. Unlike traditional KD, which relies on a single fully trained teacher model, it distributes the knowledge transfer across multiple stages and prevents reliance on a single high-capacity teacher. This is what enables the framework to effectively balance weak supervision and knowledge refinement and ensure that models trained on small-labeled subsets can still achieve competitive performance against fully supervised counterparts. Our experiments indicate that a single backbone trained on 30\% of the dataset (see Table \ref{tab:backbone_comparison}) is outperformed by the proposed chain, with the KD chain achieving an overall accuracy gain of (+23.42\%). Furthermore, the framework incorporates adaptive temperature scaling, which dynamically adjusts the softness of probability distributions throughout training to ensure a better transition from soft to sharp predictions. 

The evaluation of our model on multiple medical imaging datasets confirms its reliable generalizability across different imaging modalities. Particularly, our framework achieves accuracies of 92.61\%, 97.88\%, 88.92\%, 95.94\%, 87.70\%, and 97.18\% on the Datos, Eardrum, OtoMatch, EardrumDS, Brain Tumor, and Lung \& Colon Cancer datasets, respectively; with recall scores of 92.59\%, 97.90\%, 88.94\%, 95.94\%, 87.71\%, and 97.18\% on the respective datasets. Comparison analysis reveals that our proposed framework competes closely against most of the SOTA supervised methods, which require large-scale labeled datasets. WeCKD, in contrast, eliminates the need for a strong pretrained teacher and extensive labeled data. Additionally, it does not require any image preprocessing or quality enhancement techniques, unlike other studies that further showcase its real-world applicability.

Although the model proves to be a strong alternative to fully supervised learning, certain limitations remain that open avenues for future exploration. One key aspect is that this study was conducted solely on medical image classification. Additionally, we primarily focused on using CNN-based architectures as the backbone models. Thus, in future research, we would extend the framework to broader computer vision domains beyond medical imaging classification tasks. While our current framework works in a weakly-supervised setting with minimal labeled data, our future work direction is to develop a fully unsupervised method to eliminate reliance on annotations. In addition, alternative VIT-based architectures with their hybrid variants will be explored, which might provide even better results.
\vspace{-10pt}

\section{Conclusion} \label{conclusion}
This study introduced a novel weakly-supervised chain-based knowledge distillation framework designed for medical image classification. Unlike conventional fully supervised models that require extensive labeled datasets, WeCKD progressively refines knowledge through a structured multistage distillation chain, which allowed the framework to learn from only 30\% of the labeled data, each model taking only 10\% of the original datasets. while preserving high classification accuracy. Experimental results in multiple medical imaging datasets, including otoscopic, MRI, and microscopic images, demonstrate that the model achieves a performance comparable to that of fully supervised models. The framework effectively balances direct supervision and knowledge transfer and ensures stable generalization without requiring additional preprocessing or image enhancement techniques. A key advantage of the model is its ability to maintain knowledge consistency across multiple distillation stages and address the risk propagation errors that are commonly associated with sequential KD methods. Additionally, its structured learning approach allows for improved feature representation, which makes it well-suited for medical AI applications where labeled data are scarce. 
\vspace{-10pt}

\section*{Declarations}
\noindent
\textbf{Conflict of interests:} The authors declare no financial conflicts of interest that could have influenced this work.\\
\textbf{Ethics approval and consent to participate:} No additional ethics approval or consent was required.\\
\textbf{Data availability:} The study uses six publicly available datasets: (1) Datos \cite{Viscaino2020}, (2) Eardrum \cite{POLAT2021}, (3) OtoMatch \cite{camalan_2019_3595567}, (4) EardrumDS \cite{basaran2022eardrum}, (5) Brain Tumor dataset \cite{nickparvar2021brain}, and (6) Lung and Colon Cancer dataset \cite{dey2022lung}.


\end{document}